\documentclass[preprint,review,3p,12pt]{elsarticle}

\usepackage{amssymb}
\usepackage{soul} 
\usepackage{graphicx}
\usepackage{multicol}
\usepackage{multirow}
\usepackage{subfigure}
\usepackage{colortbl}
\usepackage{amsfonts}
\usepackage{amssymb}
\usepackage{amsmath}
\usepackage{bbding}
\usepackage{threeparttable}
\usepackage{algorithm}
\usepackage{algorithmic}
\usepackage{array}
\usepackage{color, xcolor}
\usepackage{balance}

\journal{Pattern Recognition}

\begin{document}

\begin{frontmatter}


\title{Exploiting the Potential Supervision Information of Clean Samples in Partial Label Learning}



\author[a]{Guangtai Wang}
\author[a]{Chi-Man Vong\corref{cor1}}
\author[b]{Jintao Huang}
\address[a]{Department of Computer and Information Science, University of
Macau}
\address[b]{Department of Computer Science, Hong Kong Baptist University}
\cortext[cor1]{Corresponding author, E-mail:cmvong@um.edu.mo.}
\begin{abstract}
Diminishing the impact of false-positive labels is critical for conducting disambiguation in partial label learning. However, the existing disambiguation strategies mainly focus on exploiting the characteristics of individual partial label instances while neglecting the strong supervision information of clean samples randomly lying in the datasets. In this work, we show that clean samples can be collected to offer guidance and enhance the confidence of the most possible candidates. Motivated by the manner of the \textit{differentiable count loss} strategy and \textit{the K-Nearest-Neighbor} algorithm, we proposed a new calibration strategy called \textbf{CleanSE}. Specifically, we attribute the most reliable candidates with higher significance under the assumption that for each clean sample, if its label is one of the candidates of its nearest neighbor in the representation space, it is more likely to be the ground truth of its neighbor. Moreover, clean samples offer help in characterizing the sample distributions by restricting the label counts of each label to a specific interval. Extensive experiments on 3 synthetic benchmarks and 5 real-world PLL datasets showed this calibration strategy can be applied to most of the state-of-the-art PLL methods as well as enhance their performance.
\end{abstract}


\begin{highlights}
\item A new perspective that refining clean samples helps disambiguate PLL
\item A novel reweighting method to exploit the external contact between clean and partial samples
\item A trackable way to draw the global distribution interval of samples
\end{highlights}


\end{frontmatter}


\section{Introduction}
Artificial intelligence strategies have been widely applied in the manufacturing and commercial fields due to their outstanding performance. However, these approaches rely on accurate and complete supervision information, which is impractical in many scenarios on account of high annotation costs, such as video annotation \cite{videoannotation}. Some tasks with special requirements also need data with incomplete labels, like privacy-sensitive data \cite{privacy}. Partial label learning \cite{PLLsurvey}occurred as one potential solution for tasks with difficulty obtaining accurate labels. 

Partial label learning (PLL) attributes each instance with a bag of candidates, in which the ground truth lies in the candidate label set. Since PLL has the advantage of low annotation cost and requirements of accurate labels, it represents superiority in intricate applications such as multimedia content analysis, web mining, and natural language processing. Formally speaking, considering a set of instances $\mathbf{X}=\{x_1,x_2,...,x_n\}\in \mathbb{R}^{d\times n}$, the label space is denoted as $\mathbf{Y}=\{y_1,y_2,...,y_m\} \in \mathbb{R}^{m\times 1}$, PLL aims to train a classifier $\textit{f}:\mathbf{X} \xrightarrow{}\mathbf{Y}$ from the weakly supervised dataset $D=\{\mathbf{X},\mathbf{S}\},\mathbf{S} \subset \mathbf{Y}$ to predict the labels of the new coming instances accurately. There are three main types of processing strategies: (1) disambiguation strategy: Methods that exploit the accurate label by exploring the confidence of each label related to the unlabeled instances. (2) transformation strategy: Methods that aim to transform PLL into other mature paradigms, such as binary learning \cite{binary}, dictionary learning \cite{dictionary}, and graph matching \cite{graph}. (3) Theory-oriented strategy: Methods concentrated on the theoretical perspective of PLL \cite{progressive}, \cite{robustness}. However, these methods neglect one potential supervision information: When the candidate label set contains only one label, the label included is the ground truth, which means this instance is accurately labelled. Thus, this setting serves
as our primary motivation for believing that extracting the concealed information of clean samples can guide the model in the correct way. 

Notably, the clean samples are randomly distributed among all data, which serves as an obstacle to effectively utilizing the strong supervision information. Locally, the ground truth labels of partial samples are unclear, which makes it hard to generate a relationship with the clean ones under certain prompts. Globally, clean data always occupies a small part of the dataset, making it hard to reveal the label distribution since most partial samples can offer limited, accurate information. As a result, the PLL paradigm is sensitive to the false positive candidate's labels. These two factors hinder existing PLL approaches from evolving under certain precise knowledge or specific guidance of clean samples.

Therefore, we proposed a method called \textbf{Clean} \textbf{S}ample \textbf{E}xploiting (\textbf{CleanSE}) to mitigate these aforementioned problems. To create a relationship between clean and partial samples, motivated by the data-centric perspective of candidate pruning in \cite{CLSP}, we applied a k-NN algorithm to select the most reliable labels in the candidates for the partial samples. We assume that the nearest pairs are more likely to have the same label, i.e., the labels of clean samples have a high probability of being the true label for their closest neighbours. Globally, the number of clean samples offers a minimum boundary for each label. Thus, we can roughly draw the distribution interval by treating the weak supervision as a constraint. Correspondingly, we obtained the probabilities of different label counts in a tractable computation way, which can help the model describe the distribution of whole datasets in an exact way. 

Our main contributions are listed as follows:
\begin{itemize}
\item[$\bullet$] \textbf{A new perspective that refining clean samples helps disambiguate PLL}. 

By excavating the inner structure of PLL, we can find that PLL is a supervised-and-weakly supervised problem when facing the whole situation of the dataset. Typically, the small number of clean samples can offer prompts for the model training with accurate guidance.
\item[$\bullet$]  \textbf{A novel reweighting method to exploit the external contact between clean and partial samples.} 

The reweighting measure is closely connected to the label inconsistency and consistency between the clean sample and partial samples. It further integrates the k-NN algorithm to distinguish the most similar clean-partial pairs that attach higher confidence to the common label.
\item[$\bullet$]  \textbf{A trackable way to draw the global distribution interval of samples.}

The distribution situation of all labels can be calculated by counting the positive labels of each class; we showed that there are both maximum and minimum bounds. The probability of the distribution interval can be computed tractably, which also serves as an objective for optimizing.

\item[$\bullet$]  \textbf{Therotical analysis and experiment validation are conducted.}

Theoretical analysis about the error bound explored the probable approximate correctness of the reweighting process. Estimating the distribution of extreme situations by countless also offers a potential solution for large datasets. Besides, experimental results demonstrate the effectiveness on both benchmark and real datasets.
\end{itemize}
\section{Related Works}
\subsection{Partial Label Learning} 

Partial label learning is a representative paradigm in relation to weakly supervised learning. The explicit supervision information is the ground truth concealed in the candidate set. In other words, the labels in the non-candidate set are all false labels. Based on this setting, many disambiguation methods were derived. GLMNN-PLL \cite{GLMNN-PLL} controls the disambiguation margin between similar pairs and different pairs by calculating the distances between the feature spaces. PLGP \cite{PLGP} removes the ambiguous information of labels by fitting a gaussian probabilistic kernel algorithm to observe the feature space. Unlike the aforementioned algorithms, which treat all the labels in candidates equally, SP-PLL \cite{SP-PLL} ranks the priorities of candidate labels during each learning iteration. PL-PIE \cite{PL-PIE} utilizes prior information of label distribution to determine the positive and negative labels. 

There is also a plethora of deep learning methods applied to PLL. \cite{cc-rc} develops risk-consistent (RC) and classifier-consistent (CC) methods which can be effectively utilized in deep networks under specifically designed entropy loss functions. PRODEN \cite{PRODEN} progressively updated the deep networks by process the weakly-supervised data in mini-batches. KMT-PLL \cite{KMT-PLL}claimed that the K-means cross-attention transformer could help exploit the instance-to-label relevance and a weighted cross-entropy function was adopted to train a superior classifier. PiCO+ \cite{pico+} proposed a contrastive prototype-based label disambiguation mechanism that integrates distance-based clean sample detection methods.

There are also researchers who focus on the extensions of PLL under different scenarios, such as multi-label PLL \cite{multilabelPLL}, multi-view multi-label PLL \cite{multiviewPLL}, semi-supervised PLL \cite{semi-PLL}, imbalanced PLL \cite{imbalancedPLL}, noise PLL \cite{noisePLL}, etc. Most of the previous methods have typically developed solutions under the explicit supervision that the ground truth lies in the candidates, but few researchers have concentrated on the potential clean samples hidden in the partial samples. Our previous work PLKD \cite{PLKD} focuses on using an extra knowledge distillation framework to implicitly transfer the information of clean samples to guide partial samples. Thus, we believe that new techniques that can effectively utilize these samples in an explicit way will help train an excellent classifier under weak supervision. 
\section{Methodology}
In this section, we provide explanations of CleanSE in detail. We first demonstrate the formal setting of PLL. Then, we conduct a reweighting method by utilizing k-NN to measure the relationship between clean samples and partial samples. Besides, we also proposed using a trackable label count loss as a constraint to depict the global distribution of the PLL dataset. Finally, we also show the pseudo-code to present details of the implementation. 
\subsection{Preliminaries}
\begin{figure}[tpbh]
    \centering
    \includegraphics[width=0.9\textwidth]{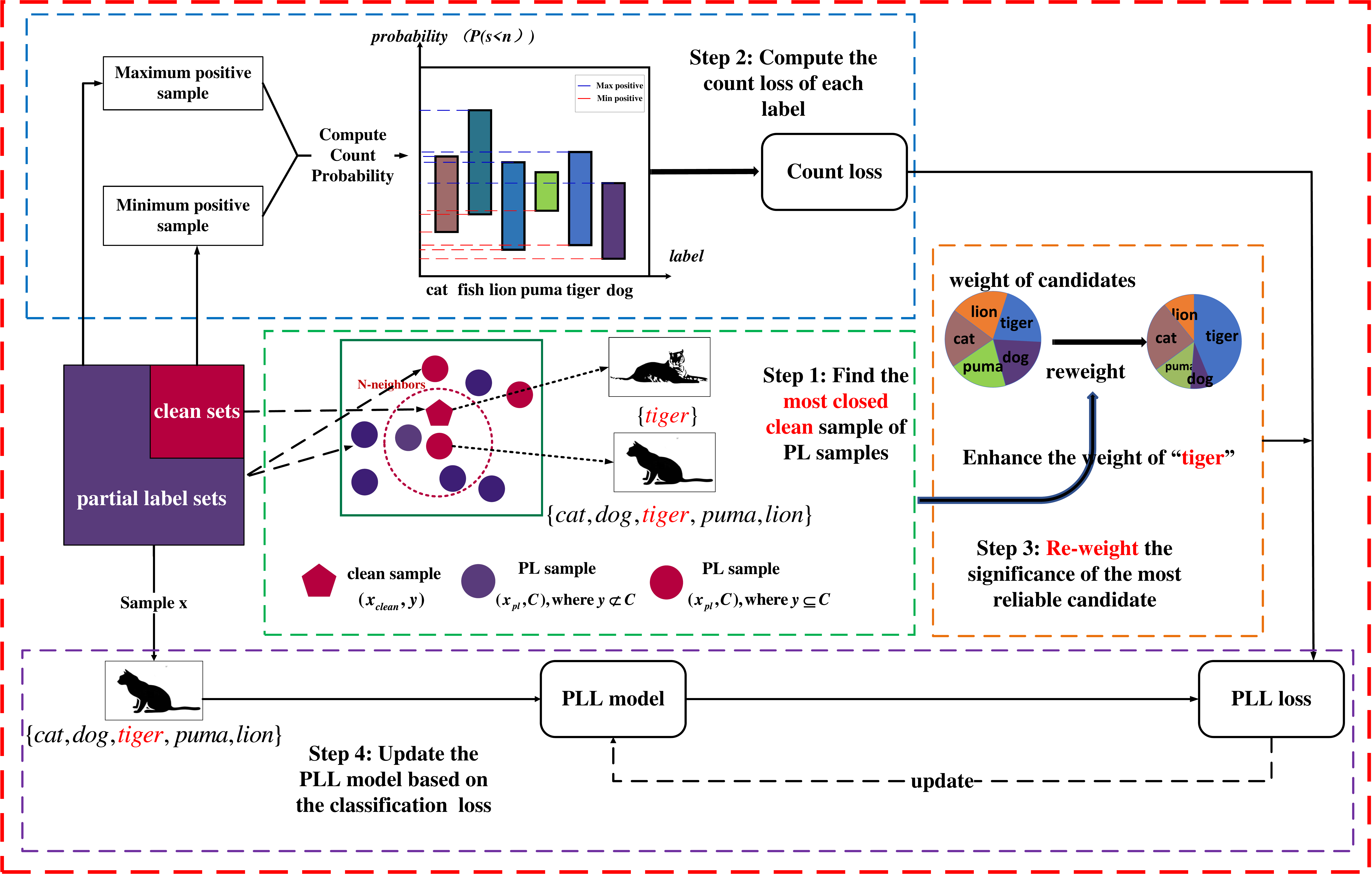}
    \caption{Training procedure of \textbf{CleanSE}. (1) The final loss function ($\mathcal{R}_{pll}$) consists of reweighting loss ($\mathcal{R}_{l}$), count loss ($\mathcal{R}_{g}$). (2) The soft labels are specifically constructed for partial labels.}
    \label{CleanSE}
\end{figure}

\subsubsection{Partial Label Learning} 
The setting of partial label learning can be formulated as follows. Given partial label dataset $D=\{\mathbf{X},\mathbf{S}\}=\{(x_i,s_i)\}_{i=1}^n$, individual instance $x_i \in \mathbf{X}$ and its corresponding candidate label set $s_i \subset \mathbf{Y}$. The basic assumption of PLL is that the true label $\bar{y}_i$ is in the candidate set $s_i$ as well as the non-candidate labels are all false. Thus, the loss function of PLL can be denoted as:
\begin{align}
\mathbb{R}&=\mathbb{E}_{(\mathbf{X},\mathbf{S})}[\mathcal{L}(\textit{f}(\mathbf{X}),\mathbf{S})]\notag\\ &=-\frac{1}{n}\sum \limits_{i=1}^n \sum \limits_{y_j\in s_i}y_j\times log(\textit{P}(x_i))\notag\\
&=-\frac{1}{n}\sum \limits_{i=1}^n \sum \limits_{y_j\in s_i}y_j\times log(\frac{exp(\textit{f}(x_i)^k)}{\sum \limits_{k=1}^mexp(\textit{f}(x_i)^k)})\notag\tag{1}
\label{1}
\end{align}

Where $\mathbb{E}$ denotes the expectation, $P(\cdot)$ demonstrates the probability, $\textit{f}(\cdot)$ represents the trained multi-class classifier, $\textit{f}(x_i)^k$ is the prediction of $k$-$th$ label trained with instance $x_i$, and $y_j$ is the $j$-$th$ label. The Cross-Entropy function is adopted, thus a Softmax activation function was employed to obtain the probability of labels. 
\subsection{Instance Reweighting}
It clearly shows that in Eq.(\ref{1}), all the candidates are equally treated, but this average-based method will fall into a dilemma that the model performance will be drastically hampered once the candidates' number is enormous, the disambiguation process is sensitive to the false positive labels \cite{PLLsurvey}.
However, the concealed ground truth can not be explicitly disclosed due to the setup of PLL. A compromised way is to attach a higher significance to the most reliable candidate for individual instances.

For each dataset satisfied the label clusterability regulation: If the features are drawn from the same feature space, the nearby features have a high probability of belonging to the same true class \cite{CLSP}, \cite{KNNSAMPLE}. To attach different significances to the candidates, we construct a transition matrix to show the different importance of labels. To this end, we can have the reweighted loss function as follows:
\begin{align}
\mathcal{R}_l&=\mathbb{E}_{(\mathbf{X},\mathbf{S})}[K\times\mathcal{L}(\textit{f}(\mathbf{X}),\mathbf{S})]\notag\\ &=-\frac{1}{n}\sum \limits_{i=1}^n \sum \limits_{j=1}^mk_{ij}y_j\times log(\textit{P}(x_i))\notag\\
&=-\frac{1}{n}\sum \limits_{i=1}^n \sum \limits_{y_j\in s_i}k_{ij}y_j\times log(\frac{exp(\textit{f}(x_i)^k)}{\sum \limits_{k=1}^mexp(\textit{f}(x_i)^k)})\notag\\&=-\frac{1}{n}\sum \limits_{i=1}^nk_{ij}\mathbb{I}(y_j\in s_i)log(\frac{exp(\textit{f}(x_i)^k)}{\sum \limits_{k=1}^mexp(\textit{f}(x_i)^k)})\tag{2}
\label{2}
\end{align}
where $K$ is the weight matrix and $k_{ij}$ denotes is weight of $j$-$th$ label in $i$-$th$ instance. $\mathbb{I} (\cdot)$ represents a indicator function.

Intuitively, the label of the clean sample is most likely to be the label of its k-NN instances. For the PLL dataset, we have $X=\{X_c \cup X_p\}$, provided that the K-NN of individual instance $x_i$ is $knn(x_i)=\{x_1,x_2,...,x_k\}$, the enhanced label $\hat{y}_i$ has multiple situations:\\
(1) $x_i$ is clean sample:

$\hat{y}_i=s_i$, where $s_i$ is the candidate set of $x_i$.\\
(2) $x_i$ is partial sample, $knn(x_i) \cap X_c=X_{and}$:
\begin{equation}
    \hat{y}_i=\begin{cases}s_k,&\text{ if }x_k \in X_{and}\text{ }\&\text{ }s_k \in s_i\\y_v,&\text{ if }x_k \in X_{and}\text{ }\&\text{ }s_k \notin s_i     
    \end{cases}\tag{3}
    \label{3}
\end{equation}\\

Here, we use a voting procedure to select the most reliable labels. By counting the frequency of each label in set $knn(x_i)$, we choose the most frequently appearing label $y_v$ as the enhanced label.\\
(3) $x_i$ is partial sample, $knn(x_i) \cap X_c=\emptyset$:

Similar to case (2), we also adopt the voting procedure to obtain the most reliable label $y_v$ as the enhanced label.

Based on the previous information, we can get the reweight matrix K as follows:
\begin{equation}
    k_{ij}=\begin{cases}0,&\text{ if }y_j \notin s_i\\ 1,&\text{ if }y_j \in s_i\text{ }\&\text{ }y_j=\hat{y}_i\\ T&\text{ if }y_j=\hat{y}_i   
    \end{cases}\tag{4}
    \label{4}
\end{equation}\\
where $T\geq 1$ denotes the temperature.\\
\textbf{Theorem 1.} \textit{Provided the loss function $\mathcal{L}$ is bounded by M, i.e., $M=sup_{x \in 
 \textbf{X}, f \in \mathcal{F}, y\in \textbf{Y}}\mathcal{L}(f(x)^k,y)$, and it is $\rho ^{'}\text{-}Lipschitz$ with respect to $f(x)(0<\rho< \infty)$ for all $y\in\textbf{Y}$, for any $\delta >0$, the following inequality exists with probability $1-\delta$:}
\begin{equation}
 \mathcal{R}_l(\hat{f_l})-\mathcal{R}_l(f^*) \leq 4\sqrt{2}\rho\sum^k_{y=1}\mathfrak{R}_n(\mathcal{H}_y)+(2T+m-2)\sqrt{\frac{log\frac{2}{\rho}}{2n}}\notag
\end{equation}
\textit{Proof. } Since the proof process is similar to that in \cite{cc-rc}, we just list the key points to show the procedure. First, we define the reweight loss function space as $\mathcal{G}_l$. The expected Rademacher complexity of $\mathcal{G}_l$ is shown as:
\begin{equation}
 \tilde{\mathfrak{R}_l}(\mathcal{G}_l)=\mathbb{E}_{\tilde{p}(X,S)}\mathbb{E}_{\sigma}[\mathop{sup}\limits_{g\in\mathcal{G}_l}\frac{1}{n}\sum^n_{i=1}\sigma_ig(x_i,s_i)]\tag{5}
\end{equation}
Suppose $D$ and $D^{'}$ be two samples differing by exactly one point, which is $(x_i,s_i)\in D$ and $(x_i^{'},s_i^{'})\in D^{'}$, we have:
\begin{align}
       \mathcal{R}_l(D) -\mathcal{R}_l(\mathbb{D^{'}}) &\leq \mathop{sup}\limits_{f\in\mathcal{F}}\frac{|k_i-k_i^{'}|M}{n}\notag\\&=\frac{(2T+m-2)M}{n}\tag{6} 
\end{align}
Based on McDiarmid’s inequality, we can get:

\begin{align}
    \mathop{sup}\limits_{f\in \mathcal{F}}\mathcal{R}_l(f)-&\hat{\mathcal{R}}_l(f)\leq \mathbb{E}[\mathop{sup}\limits_{f\in \mathcal{F}}\mathcal{R}_l(f)-\hat{\mathcal{R}}_l(f)]\notag\\  &+ (2T+m-2)\sqrt{\frac{log\frac{2}{\rho}}{2n}}\tag{7}
\end{align}
which exists with probability at least $1-\delta/2$ for any $\delta>0$. Using the trick mentioned in \cite{MLproof}, we have $\mathbb{E}[\mathop{sup}\limits_{f\in \mathcal{F}}\mathcal{R}_l(f)-\hat{\mathcal{R}}_l(f)]\leq2\tilde{\mathfrak{R}}_n(\mathcal{G}_l)$. Furthermore, the following inequality exists with probability at least $1-\delta$:
\begin{align}
    |\mathop{sup}\limits_{f\in \mathcal{F}}\mathcal{R}_l(f)-&\hat{\mathcal{R}}_l(f)|\leq2\tilde{\mathfrak{R}}_n(\mathcal{G}_l)+(2T+m-2)\sqrt{\frac{log\frac{2}{\rho}}{2n}}\tag{8}
\end{align}
For all $y\in Y$, based on the Rademacher vector contraction inequality, we can get $\tilde{\mathfrak{R}}_n(\mathcal{G}_l)\leq\sqrt{2}\rho\sum_{y=1}^m\mathfrak{R}_n(\mathcal{H}_y)$. Thus \textbf{Theorem 1.} is proved.
\subsection{Countloss} 
Recall that we can denote the partial label dataset as $D=\{D_c, D_p\}$ from the perspective of regarding it as a supervised-and-weakly supervised problem. Since the non-candidate label has no probability to be the ground truth, for individual label $y_j$, one situation needs to be observed: The number of instances labeled as $y_j$,i.e., $N_j$ lies in a certain interval:
\begin{equation}
    N_c^j \leq N_j \leq N_p^j+N_c^j \tag{9}
    \label{9}
\end{equation}
The label frequency can be obtained by counting the candidates, $N_c^j$, and $N_p^j$ are the number of clean and partial samples labeled as $y_j$, the value of  $j$-$th$ label in $i$-$th$ instance is denoted as $y_j^i$. Our objective function is to minimize the entropy:
\begin{align}
     \mathcal{R}_g&=-\sum \limits_{j=1}^mp(N_c^j \leq \sum \limits_{i=1}^ny_j^i \leq N_p^j+N_c^j)\notag\\&\times\text{log }p(N_c^j \leq \sum \limits_{i=1}^ny_j^i \leq N_p^j+N_c^j)\tag{10}
     \label{10}
\end{align}
For Eq.(\ref{10}), we denote events $\sum \limits_{i=1}^ny_j^i =k$ as $\theta_n^k$, since the dataset is fixed, which means that case $A(\theta_n^k)$ and case $B(\theta_n^s)$ are pairwise mutually exclusive events if $k\neq s$, we end up with the probability by follows:
\begin{equation}
     p(N_c^j \leq \sum \limits_{i=1}^ny_j^i \leq N_p^j+N_c^j)=\sum \limits_{k=N_c^j}^{N_p^j+N_c^j}p(\theta_n^k)\tag{11}
     \label{11}
\end{equation}
The count label loss can be trackably computed based on the partition theorem in \cite{simple}, \cite{countloss}. To obtain the count probability $p(\theta_n^k)$, we have:

\begin{align}
        p(\theta_n^k)=&p(\theta_{n-1}^{k-1}) \times p(y_j^n=1)\notag+p(\theta_{n-1}^k) \times p(y_j^n=0) 
        \tag{12}
        \label{12}
\end{align}

Assuming the partial samples are independently drawn from the distribution $p(x, Y)$, the partial data generation process of individual instance $x_i$ whose ground truth is $y_j$ follows a binomial distribution $ s_i\sim (m-1,0.5)$. We can imagine when the number of $j$-$th$ label the whole dataset $N_j\rightarrow\infty$, the number of clean sample $N_c^j\rightarrow\frac{N_j}{2^{m-2}}$. Provided the number of dataset is $N$, for the whole dataset, the average partial label number is $\frac{m+1}{2}$, thus $N_p^j+N_c^j=\frac{N(m+1)}{2}\times (N-N_j)+N_j$. Here we can estimate the portion of each label $p_j=\frac{N_j}{N}$in the dataset by counting the frequency of clean samples and partial samples, the number of $j$-$th$ label $N_j=2^{m-2}N_c^j$, and the number of instances of whole datasets can be denoted as:
\begin{align}
N&={\frac{N_j+\sqrt{N_j^2+\frac{8}{m+1}(N^j_p+N_c^j-N_j)}}{2}}\notag\\&=\sqrt{(2^{m-3}N_c^j)^2+\frac{2}{m+1}(N_p^j+N_c^j+2^(m-2)N_c^j)}\notag\\&+2^{m-3}N_c^j\notag\\&=\sqrt{4^{m-3}(N_c^j)^2+\frac{2}{m+1}(N_p^j+N_c^j+2^(m-2)N_c^j}\notag\\&2^{m-3}N_c^j\tag{13}
\end{align}
Thus for large datasets, we can also minimize the KL divergence between the distribution of predicted label sum and label portion of the instances, by:
\begin{equation}
   \mathcal{R}_g=\sum_{j=1}^m\mathbb{D}_{KL}(p_j||p(\sum_{i=1}^n=N_j))\tag{14}
\end{equation}

\subsection{CleanSE Loss}
In the previous section, we showed two terms for utilizing clean samples to enhance the model performance. So, the final objective function of CleanSE is shown as:
\begin{equation}
    \mathcal{R}_{pll}=\mathcal{R}_c+\lambda\mathcal{R}_l\tag{15}
    \label{15}
\end{equation}
where $\lambda$ is a hyperparameter. 

Fig. \ref{CleanSE} depicts the whole framework of CleanSE, and the pseudo-code is presented in Algorithm \ref{CleanSR_ALGO}.
\begin{algorithm}[tph]
    \caption{Training of PLKD}
    \label{CleanSR_ALGO}
    	\renewcommand{\algorithmicrequire}{\textbf{Input:}}
	  \renewcommand{\algorithmicensure}{\textbf{Output:}}
        \begin{algorithmic}[1]
        \REQUIRE{PLL dataset $D=\{D_c \cup D_p\}=\{x_i,s_i\}_{i=1}^n$; training epoch $E$ and iteration $I$; count loss regularization hyperparameter $\lambda$; and reweight temperature T}
        \STATE{Initialize the model $\theta$} 
        \FOR{{$t=1,...,E$}}
        \STATE{Shuffle the partial label dataset $D$ into $B$ mini-batches;}
        \FOR{{$I=1,...,B$}}
        \STATE{Get the reweight matrix $K$ based on Eq.(\ref{3}, \ref{4});} 
        \STATE{Get reweighted loss $\mathcal{R}_l$ based on Eq.(\ref{2})}  
        \STATE{Calculate the number of clean and partial samples in the mini-batch, obtain $N_c=\{N_c^1,...,N_c^m\}$ and $N_p=\{N_p^1,...,N_p^m\}$} 
        \STATE{Get the count probability $p(\sum \limits_{i=1}^ny_j^i=k)$ based on Eq.(\ref{12});}
        \STATE{Get the count loss $\mathcal{R}_{g}$ based on Eq.(\ref{11}),Eq.(\ref{10});} 
        \STATE{Get PLL loss $\mathcal{R}_{pll}$ based on Eq.(\ref{15});} 
        \STATE{Update $\theta$ by minimizing the PLL loss $\mathcal{R}_{pll}$ with forward computation and back-propagation;}
        \ENDFOR
        \ENDFOR
        \ENSURE{model $\theta$}
    \end{algorithmic}
\end{algorithm}
\begin{table}[htbp]
\caption{Characteristics of the Mentioned Datasets}
\resizebox{\linewidth}{!}{
\begin{tabular}{llllll}
\hline
Datasets&Instance number&Feature&Classes&Averge PLs&CS-Rate\\\hline
MNIST&70,000&28x28&10&4.9992&0.93\%\\
Fashion-MNIST&70,000&28x28&10&5.0078&0.94\%\\
Kuzushiji-MNIST&70,000&28x28&10&5.0028&1.03\%\\
CIFAR-10&60,000&32x32&10&4.9968&0.95\%\\\hline\hline
Lost (Lo)&1,122&108&16&2.2300&6.24\%\\
BirdSong (Bs)&4,998&38&13&2.1800&33.22\%\\
MSRCv2 (Ms)&1,758&48&23&3.1600&7.54\%\\
Soccer Player (Sp)&17,472&279&171&2.0900&30.45\%\\
Yahoo! News (Yn)&22,991&163&219&1.9100&28.99\% \\
\hline\hline
\end{tabular}}
\label{datasets}
\end{table}
\begin{table*}[htbp]\normalsize
\centering
\caption{Prediction Performance on MNIST Dataset}
\resizebox{\linewidth}{!}{
\begin{tabular}{l|lllll}
\cline{1-6}
\multicolumn{1}{c}{Batch Size}& \multicolumn{1}{c}{16}& \multicolumn{1}{c}{32}& \multicolumn{1}{c}{64}& \multicolumn{1}{c}{128}& \multicolumn{1}{c}{256}\\ \cline{1-6}
\multicolumn{1}{c}{Methods}         & \multicolumn{1}{c}{Accuracy} & \multicolumn{1}{c}{Accuracy} & \multicolumn{1}{c}{Accuracy}  & \multicolumn{1}{c}{Accuracy} & \multicolumn{1}{c}{Accuracy}  \\ \hline\hline
CC & \multicolumn{1}{l|}{97.31 ± 0.25\%(4)}             & \multicolumn{1}{l|}{97.50 ± 0.29\%(4)}              & \multicolumn{1}{l|}{97.76 ± 0.26\%(3)}                & \multicolumn{1}{l|}{97.81 ± 0.18\%(3)}              & \multicolumn{1}{l}{97.86 ± 0.16\%(4)}                                                        \\ 
RC & \multicolumn{1}{l|}{97.43 ± 0.13\%(2)}             & \multicolumn{1}{l|}{97.52 ± 0.11\%(3)}              & \multicolumn{1}{l|}{97.86 ± 0.13\%(2)}                & \multicolumn{1}{l|}{97.69 ± 0.42\%(4)}              & \multicolumn{1}{l}{97.94 ± 0.08\%(2)}                                                        \\
LWS & \multicolumn{1}{l|}{97.36 ± 0.09\%(3)}             & \multicolumn{1}{l|}{97.56 ± 0.11\%(2)}              & \multicolumn{1}{l|}{97.75 ± 0.19\%(4)}                & \multicolumn{1}{l|}{97.92 ± 0.07\%(2)}              & \multicolumn{1}{l}{97.89 ± 0.12\%(3)}                                                        \\
CAVL & \multicolumn{1}{l|}{96.14 ± 0.22\%(5)}             & \multicolumn{1}{l|}{96.76 ± 0.11\%(5)}              & \multicolumn{1}{l|}{96.91 ± 0.11\%(5)}                & \multicolumn{1}{l|}{97.05 ± 0.08\%(5)}              & \multicolumn{1}{l}{97.13 ± 0.09\%(5)}                                                        \\\hline\hline
SCARCE & \multicolumn{1}{l|}{90.37 ± 0.57\%(6)}             & \multicolumn{1}{l|}{90.72 ± 0.51\%(6)}              & \multicolumn{1}{l|}{91.88 ± 0.35\%(6)}                & \multicolumn{1}{l|}{94.25 ± 0.17\%(6)}              & \multicolumn{1}{l}{95.19 ± 0.15\%(6)}\\
\hline\hline
PL-CGR&\multicolumn{1}{c}{ \text{-}}&\multicolumn{1}{c}{ \text{-}}&\multicolumn{1}{c}{ \text{-}}&\multicolumn{1}{c}{ \text{-}}&\multicolumn{1}{c}{ \text{-}}\\
DPCLS&\multicolumn{1}{c}{ \text{-}}&\multicolumn{1}{c}{ \text{-}}&\multicolumn{1}{c}{ \text{-}}&\multicolumn{1}{c}{ \text{-}}&\multicolumn{1}{c}{ \text{-}}\\\hline\hline
Ours & \multicolumn{1}{l|}{\textbf{97.74 ± 0.09\%(1)}}             & \multicolumn{1}{l|}{\textbf{97.90 ± 0.15\%(1)}}              & \multicolumn{1}{l|}{\textbf{97.97 ± 0.09\%(1)}}                & \multicolumn{1}{l|}{\textbf{98.07 ± 0.07\%(1)}}              & \multicolumn{1}{l}{\textbf{98.19 ± 0.10\%(1)}}\\\hline\hline
\end{tabular}}
\label{mnist_test}
\end{table*}
\begin{table*}[htbp]\normalsize
\centering
\caption{Prediction Performance on KMNIST Dataset}
\resizebox{\linewidth}{!}{
\begin{tabular}{l|lllll}
\cline{1-6}
\multicolumn{1}{c}{Batch Size}& \multicolumn{1}{c}{16}& \multicolumn{1}{c}{32}& \multicolumn{1}{c}{64}& \multicolumn{1}{c}{128}& \multicolumn{1}{c}{256}\\ \cline{1-6}
\multicolumn{1}{c}{Methods}         & \multicolumn{1}{c}{Accuracy} & \multicolumn{1}{c}{Accuracy} & \multicolumn{1}{c}{Accuracy}  & \multicolumn{1}{c}{Accuracy} & \multicolumn{1}{c}{Accuracy}  \\ \hline\hline
CC & \multicolumn{1}{l|}{87.25 ± 0.50\%(3)}             & \multicolumn{1}{l|}{87.81 ± 0.29\%(4)}              & \multicolumn{1}{l|}{88.72 ± 0.45\%(3)}                & \multicolumn{1}{l|}{88.74 ± 1.28\%(4)}              & \multicolumn{1}{l}{88.62 ± 0.67\%(3)}                                                        \\ 
RC & \multicolumn{1}{l|}{87.14 ± 0.47\%(4)}             & \multicolumn{1}{l|}{88.34 ± 0.44\%(2)}              & \multicolumn{1}{l|}{88.69 ± 0.37\%(4)}                & \multicolumn{1}{l|}{89.44 ± 0.39\%(3)}              & \multicolumn{1}{l}{89.76 ± 0.35\%(3)}                                                        \\
LWS & \multicolumn{1}{l|}{87.47 ± 0.25\%(2)}             & \multicolumn{1}{l|}{88.24 ± 0.33\%(3)}              & \multicolumn{1}{l|}{88.93 ± 0.42\%(2)}                & \multicolumn{1}{l|}{89.81 ± 0.18\%(2)}              & \multicolumn{1}{l}{89.81 ± 0.62\%(2)}                                                        \\
CAVL & \multicolumn{1}{l|}{84.21 ± 0.49\%(5)}             & \multicolumn{1}{l|}{85.12 ± 0.45\%(5)}              & \multicolumn{1}{l|}{85.80 ± 0.43\%(5)}                & \multicolumn{1}{l|}{86.61 ± 0.41\%(5)}              & \multicolumn{1}{l}{86.63 ± 0.28\%(5)}                                                        \\
\hline\hline
SCARCE & \multicolumn{1}{l|}{66.36 ± 0.79\%(6)}             & \multicolumn{1}{l|}{67.46 ± 1.26\%(6)}              & \multicolumn{1}{l|}{72.19 ± 1.07\%(6)}                & \multicolumn{1}{l|}{76.05 ± 0.78\%(6)}              & \multicolumn{1}{l}{78.81 ± 0.67\%(6)}\\\hline\hline
PL-CGR&\multicolumn{1}{c}{ \text{-}}&\multicolumn{1}{c}{ \text{-}}&\multicolumn{1}{c}{ \text{-}}&\multicolumn{1}{c}{ \text{-}}&\multicolumn{1}{c}{ \text{-}}\\
DPCLS&\multicolumn{1}{c}{ \text{-}}&\multicolumn{1}{c}{ \text{-}}&\multicolumn{1}{c}{ \text{-}}&\multicolumn{1}{c}{ \text{-}}&\multicolumn{1}{c}{ \text{-}}\\\hline\hline
Ours & \multicolumn{1}{l|}{\textbf{88.04 ± 0.42\%(1)}}             & \multicolumn{1}{l|}{\textbf{89.37 ± 0.30\%(1)}}              & \multicolumn{1}{l|}{\textbf{90.00 ± 0.19\%(1)}}                & \multicolumn{1}{l|}{\textbf{90.49 ± 0.29\%(1)}}              & \multicolumn{1}{l}{\textbf{90.58 ± 0.38\%(1)}}\\\hline\hline
\end{tabular}}
\label{kmnist_test}
\end{table*}
\begin{table*}[htbp]\normalsize
\centering
\caption{Prediction Performance on FASHION-MNIST Dataset}
\resizebox{\linewidth}{!}{
\begin{tabular}{l|lllll}
\cline{1-6}
\multicolumn{1}{c}{Batch Size}& \multicolumn{1}{c}{16}& \multicolumn{1}{c}{32}& \multicolumn{1}{c}{64}& \multicolumn{1}{c}{128}& \multicolumn{1}{c}{256}\\ \cline{1-6}
\multicolumn{1}{c}{Methods}         & \multicolumn{1}{c}{Accuracy} & \multicolumn{1}{c}{Accuracy} & \multicolumn{1}{c}{Accuracy}  & \multicolumn{1}{c}{Accuracy} & \multicolumn{1}{c}{Accuracy}  \\ \hline\hline
CC & \multicolumn{1}{l|}{86.62 ± 0.38\%(2)}             & \multicolumn{1}{l|}{86.81 ± 0.44\%(3)}              & \multicolumn{1}{l|}{87.39 ± 0.37\%(3)}                & \multicolumn{1}{l|}{87.69 ± 0.38\%(3)}              & \multicolumn{1}{l}{87.93 ± 0.56\%(4)}                                                        \\ 
RC & \multicolumn{1}{l|}{86.20 ± 0.07\%(3)}             & \multicolumn{1}{l|}{86.95 ± 0.17\%(2)}              & \multicolumn{1}{l|}{87.40 ± 0.12\%(2)}                & \multicolumn{1}{l|}{88.08 ± 0.20\%(2)}              & \multicolumn{1}{l}{88.29 ± 0.17\%(2)}                                                        \\
LWS & \multicolumn{1}{l|}{85.20 ± 0.21\%(4)}             & \multicolumn{1}{l|}{86.35 ± 0.26\%(4)}              & \multicolumn{1}{l|}{87.06 ± 0.19\%(4)}                & \multicolumn{1}{l|}{87.55 ± 0.20\%(4)}              & \multicolumn{1}{l}{88.09 ± 0.21\%(3)}                                                        \\
CAVL & \multicolumn{1}{l|}{85.16 ± 0.42\%(5)}             & \multicolumn{1}{l|}{85.59 ± 0.37\%(5)}              & \multicolumn{1}{l|}{86.22 ± 0.14\%(5)}                & \multicolumn{1}{l|}{86.83 ± 0.18\%(5)}              & \multicolumn{1}{l}{86.91 ± 0.14\%(5)}\\
\hline\hline
SCARCE & \multicolumn{1}{l|}{81.85 ± 0.36\%(6)}             & \multicolumn{1}{l|}{82.41 ± 0.64\%(6)}              & \multicolumn{1}{l|}{84.12 ± 0.40\%(6)}                & \multicolumn{1}{l|}{84.72 ± 0.54\%(6)}         & \multicolumn{1}{l}{85.13 ± 0.37\%(6)}                                                        \\\hline\hline
PL-CGR&\multicolumn{1}{c}{ \text{-}}&\multicolumn{1}{c}{ \text{-}}&\multicolumn{1}{c}{ \text{-}}&\multicolumn{1}{c}{ \text{-}}&\multicolumn{1}{c}{ \text{-}}\\
DPCLS&\multicolumn{1}{c}{ \text{-}}&\multicolumn{1}{c}{ \text{-}}&\multicolumn{1}{c}{ \text{-}}&\multicolumn{1}{c}{ \text{-}}&\multicolumn{1}{c}{ \text{-}}\\\hline\hline
Ours & \multicolumn{1}{l|}{\textbf{87.44 ± 0.28\%(1)}}             & \multicolumn{1}{l|}{\textbf{87.97 ± 0.26\%(1)}}              & \multicolumn{1}{l|}{\textbf{88.44 ± 0.16\%(1)}}               & \multicolumn{1}{l|}{\textbf{88.66 ± 0.23\%(1)}}              & \multicolumn{1}{l}{\textbf{88.83 ± 0.16\%(1)}}\\\hline\hline
\end{tabular}}
\label{fashion_test}
\end{table*}
\begin{table*}[htbp]\normalsize
\centering
\caption{Prediction Performance on CIFAR-10 Dataset}
\resizebox{\linewidth}{!}{
\begin{tabular}{l|lllll}
\cline{1-6}
\multicolumn{1}{c}{Batch Size}& \multicolumn{1}{c}{16}& \multicolumn{1}{c}{32}& \multicolumn{1}{c}{64}& \multicolumn{1}{c}{128}& \multicolumn{1}{c}{256}\\ \cline{1-6}
\multicolumn{1}{c}{Methods}         & \multicolumn{1}{c}{Accuracy} & \multicolumn{1}{c}{Accuracy} & \multicolumn{1}{c}{Accuracy}  & \multicolumn{1}{c}{Accuracy} & \multicolumn{1}{c}{Accuracy}  \\ \hline\hline
CC & \multicolumn{1}{l|}{77.76 ± 0.44\%(4)}             & \multicolumn{1}{l|}{77.07 ± 0.35\%(4)}              & \multicolumn{1}{l|}{76.22 ± 0.34\%(3)}                & \multicolumn{1}{l|}{75.46 ± 0.55\%(3)}              & \multicolumn{1}{l}{72.59 ± 0.23\%(4)}                                                        \\ 
RC & \multicolumn{1}{l|}{78.08 ± 0.44\%(3)}             & \multicolumn{1}{l|}{77.14 ± 0.23\%(3)}              & \multicolumn{1}{l|}{74.52 ± 0.38\%(5)}                & \multicolumn{1}{l|}{74.09 ± 0.48\%(5)}              & \multicolumn{1}{l}{73.13 ± 1.10\%(3)}\\
LWS & \multicolumn{1}{l|}{78.66 ± 0.35\%(2)}             & \multicolumn{1}{l|}{77.26 ± 0.39\%(2)}              & \multicolumn{1}{l|}{76.73 ± 0.31\%(2)}                & \multicolumn{1}{l|}{75.75 ± 0.99\%(2)}             & \multicolumn{1}{l}{73.55 ± 1.78\%(2)}\\
CAVL & \multicolumn{1}{l|}{70.34 ± 0.64\%(5)}             & \multicolumn{1}{l|}{76.50 ± 0.35\%(5)}              & \multicolumn{1}{l|}{75.91 ± 0.54\%(5)}                & \multicolumn{1}{l|}{74.88 ± 0.53\%(5)}              & \multicolumn{1}{l}{71.37 ± 0.69\%(5)} \\\hline\hline
SCARCE & \multicolumn{1}{l|}{59.33 ± 0.52\%(6)}             & \multicolumn{1}{l|}{62.01 ± 0.56\%(6)}            & \multicolumn{1}{l|}{64.46 ± 0.45\%(6)}                & \multicolumn{1}{l|}{63.48 ± 0.40\%(6)}              & \multicolumn{1}{l}{60.96 ± 0.61\%(6)}\\\hline\hline
PL-CGR&\multicolumn{1}{c}{ \text{-}}&\multicolumn{1}{c}{ \text{-}}&\multicolumn{1}{c}{ \text{-}}&\multicolumn{1}{c}{ \text{-}}&\multicolumn{1}{c}{ \text{-}}\\
DPCLS&\multicolumn{1}{c}{ \text{-}}&\multicolumn{1}{c}{ \text{-}}&\multicolumn{1}{c}{ \text{-}}&\multicolumn{1}{c}{ \text{-}}&\multicolumn{1}{c}{ \text{-}}\\\hline\hline
Ours & \multicolumn{1}{l|}{\textbf{79.12 ± 0.30\%(1)}}             & \multicolumn{1}{l|}{\textbf{79.06 ± 0.20\%(1)}}              & \multicolumn{1}{l|}{\textbf{77.81 ± 0.39\%(1)}}                & \multicolumn{1}{l|}{\textbf{77.09 ± 0.59\%(1)}}              & \multicolumn{1}{l}{\textbf{74.50 ± 0.91\%(1)}}\\\hline\hline
\end{tabular}}
\label{cifar_test}
\end{table*}
\begin{table*}[htbp]\normalsize
\caption{Prediction Performance on Real-world Partial Label Dataset}
\label{real_test}
\resizebox{\linewidth}{!}{
\begin{tabular}{l|lllll}
\cline{1-6}
\multicolumn{1}{c}{Dataset}& \multicolumn{1}{c}{Lost}& \multicolumn{1}{c}{BirdSong}& \multicolumn{1}{c}{MSRCv2}& \multicolumn{1}{c}{Soccer Player} & \multicolumn{1}{c}{Yahoo! News}\\ \hline\hline
CC & \multicolumn{1}{l}{63.11 ± 1.15\%(5)}& \multicolumn{1}{l}{70.83 ±  0.48\%(4)}& \multicolumn{1}{l}{50.72 ± 0.79\%(4)}&\multicolumn{1}{l}{54.87 ± 0.38\%(4)}&\multicolumn{1}{l}{65.67 ± 0.37\%(4)}\\ 
RC & \multicolumn{1}{l}{63.21 ± 1.03\%(4)}& \multicolumn{1}{l}{72.26 ± 0.39\%(3)}&\multicolumn{1}{l}{51.14 ± 1.45\%(3)}&\multicolumn{1}{l}{55.26 ± 0.19\%(3)}&\multicolumn{1}{l}{65.89 ± 0.36\%(3)}\\ 
LWS & \multicolumn{1}{l}{61.60 ± 0.72\%(7)}& \multicolumn{1}{l}{72.32 ± 0.25\%(2)}&  \multicolumn{1}{l}{48.80 ± 0.26\%(6)}&\multicolumn{1}{l}{54.07 ± 0.13\%(5)}&\multicolumn{1}{l}{64.73 ± 0.34\%(5)}\\ 
CAVL & \multicolumn{1}{l}{61.91 ± 0.66\%(6)}& \multicolumn{1}{l}{69.77 ± 0.54\%(6)}&  \multicolumn{1}{l}{47.55 ± 1.04\%(8)}&\multicolumn{1}{l}{51.03 ± 0.43\%(7)}&\multicolumn{1}{l}{58.22 ± 0.91\%(7)}\\ \hline\hline
SCARCE & \multicolumn{1}{l}{59.55 ± 0.74\%(8)}& \multicolumn{1}{l}{69.84 ± 0.82\%(5)}&  \multicolumn{1}{l}{48.06 ± 0.74\%(7)}&\multicolumn{1}{l}{49.64 ± 0.07\%(8)}&\multicolumn{1}{l}{54.95 ± 0.12\%(8)}\\ \hline\hline
PL-CGR & \multicolumn{1}{l}{\textbf{80.59 ± 0.07\%(1)}}& \multicolumn{1}{l}{52.10 ± 0.02\%(7)}&  \multicolumn{1}{l}{\textbf{57.63 ± 0.04\%(1)}}&\multicolumn{1}{l}{56.50 ± 0.01\%(2)}&\multicolumn{1}{l}{66.46 ± 0.37\%(2)}\\ 
DPCLS & \multicolumn{1}{l}{72.28 ± 0.72\%(2)}& \multicolumn{1}{l}{49.45 ± 0.19\%(8)}&  \multicolumn{1}{l}{50.26 ± 0.21\%(5)}&\multicolumn{1}{l}{53.13 ± 0.21\%(6)}&\multicolumn{1}{l}{63.26 ± 1.10\%(6)}\\\hline\hline
Ours & \multicolumn{1}{l}{69.52 ± 1.97\%(3)}& \multicolumn{1}{l}{\textbf{75.34 ± 0.38\%(1)}}&  \multicolumn{1}{l}{55.56 ± 1.07\%(2)}&\multicolumn{1}{l}{\textbf{58.53 ± 0.30\%(1)}}&\multicolumn{1}{l}{\textbf{67.49 ± 0.23\%(1)}}\\ \hline\hline
\end{tabular}}
\end{table*}
\subsection{Complexity Analysis}
From the Algorithm and the previous sections, we need to analyze time complexity in the following 2 parts:\\
(1) Instance Reweighting: A k-NN scheme is utilized, for each training observation $x_i$ with dimension $d$ in the set, it requires $\mathcal{O}(d)$ runtime per distance computation procedure. Thus, for n samples, the distance computation will cost $\mathcal{O}(nd)$, then returning the K indices costs $\mathcal{O}(kn)$, and the reweighting process consumes $\mathcal{O}(n)$. In each training iteration, it takes $\mathcal{O}_{re}=\mathcal{O}((d+1+k)n)$ runtime to reweight all the instances.\\
(2) Count Loss: Based on the proposition \cite {simple}, the trackble computation process $p(\sum \limits_{i=1}^ny_j^i=s)$ costs $\mathcal{O}(ns)$ runtime, the sum probability will cost:
\begin{equation}
\mathcal{O}_{cl}=\mathcal{O}(\sum \limits_{N=N_c}^{N_c+N_p}Nn)=\mathcal{O}(\frac{(N_p(2N_c+N_p))}{2}n)\tag{16}
\end{equation}
From the discussion above, we have the time complexity $\mathcal{O}_{cs}=\mathcal{O}_{re}+\mathcal{O}_{cl}$.
\section{Experiments}
\subsection{Experiments Setup}
\noindent \textbf{Datasets.} The experimental evaluation was conducted across 4 benchmark datasets and 5 real-world PLL datasets. \textbf{MNIST} \cite{mnist}, \textbf{Fashion-MNIST} \cite{fmnist}, and \textbf{Kuzushiji-MNIST} \cite{kmnist} originated from the handwriting digits classification tasks, which are commonly used in PLL fields. \textbf{CIFAR-10} \cite{CIFAR10}, which consists of 32×32 colour images, is also a suitable benchmark for generating conventional partially labelled datasets. In addition, the 5 widely used real-world PLL datasets, such as \textbf{Lost} \cite{lost}, \textbf{BirdSong} \cite{birdsong}, \textbf{MSRCv2} \cite{msrcv2}, \textbf{Soccer Player} \cite{soccerplayer}, \textbf{Yahoo! News} \cite{yahoonews}, were adopted to verify the effectiveness of the proposed CleanSE. The quantitative characteristic details about these datasets are shown in Table \ref{datasets}.

\noindent \textbf{Compared Methods.} We chose 7 state-of-the-art PLL methods to compare:
\begin{itemize}
    \item \textbf{CC} and \textbf{RC} \cite{cc-rc}, \textbf{CC} is a classifier-consistency method that estimates error for all instances with a transition matrix. \textbf{RC} is a risk-consistency method for training models by estimating the prediction confidence of each instance.
    \item \textbf{LWS} \cite{lws}, it is an algorithm that offers a way to measure the risk by balancing the weight between candidate and non-candidate labels.  
    \item \textbf{CAVL} \cite{cavl}, it utilizes the maximum class activation value to exploit the true label for model training.
    \item \textbf{SCARCE} \cite{SCARCE}, it proposed an unbiased risk estimator from the perspective of positive and unlabelled learning, which treats the complementary label learning paradigm as negative-unlabeled binary classification problems.
    \item \textbf{PL-CGR} \cite{PL-CGR}, it adopts a cost-guided retraining strategy to modify and correct the disambiguation process gradually.
    \item \textbf{DPCLS} \cite{DPCLS}, it constructs similarity and dissimilarity matrices to measure the relation between samples globally.
\end{itemize}

\noindent \textbf{Implementation.} The training and test sets are randomly selected, with batch sizes systematically varied across \{16, 32, 64, 128, 256\} to investigate batch dimension effects. Empirically, the training epoch was set as 250 to make the model nearly completely converge. In the experiments of 3 MNIST datasets, we adopted a 3-layer-MLP model as the baseline for the 4 deep learning algorithms, and we used a ResNet-34 to test the performance on Cifar-10. All the hyperparameter configurations of the comparative algorithms are set as their default. In our method, the learning rate is $10^{-3}$ with $10^{-5}$ weight decay. The count loss hyperparameter $\alpha$ is $10^{-3}$. All the experiments executed through PyTorch on a NVIDIA GeForce GTX 2080.

Notably, operating the count loss procedure relies on calculating the product of probabilities of different labels. Due to the characteristics of the computer, it is extremely prone to numerical underflow during large-batch processing. To ensure numerical stability, we embed the requisite data into log space. Specifically, for a log probability $log(p)$, \textbf{log1mexp} and \textbf{logsumexp} function are used to compute $log(1-p)$ and $log(p_1+p_2)$. 
\subsection{Benchmark Test} \label{benchmark test}
In this section, we carried out a series of experiments on 4 artificial synthetic PLL datasets generated based on the generation protocol in \cite{cc-rc}. All the partial labels are drawn from a uniform probability distribution under the assumption that each partial sample is independent. 

Comparative results across benchmark algorithms are quantitatively summarized in Table \ref{mnist_test}-Table \ref{cifar_test}. The best result is highlighted in bold form, the mean accuracy and standard deviation of the last 10 epochs (mean ± std) are displayed, and we also ranked each algorithm by its performance. From the outcome in these tables, we can observe that:
\begin{itemize}
    \item CleanSE demonstrates consistent superiority over the other five algorithms on all the benchmark synthetic PLL datasets.
    \item Unlike the results in \textbf{CIFAR-10}, CleanSE's performance positively correlates with batch size expansion in \textbf{MNIST}, \textbf{KMNIST}, and \textbf{Fashion-MNIST}. Based on the result in \cite{countloss}, we can see that the effect of countless schemes will be better if the batch size is smaller. Since k-NN behaves better when the batch size is bigger, we can conclude that the count loss scheme plays a dominate role in CIFAR-10, while reweighting mechanisms prevail in the handwriting datasets. The re-weighting classification loss took a dominant position since the count loss can roughly draw the distribution interval with more weakly supervision information. 
    \item The results of the two machine learning methods, \textbf{PL-CGR} and \textbf{DPCLS}, are not listed. This is because their optimization processes are based on large matrix computation, which is easily out of memory when we operate the algorithm on our equipment. Though we didn't get their predictive performance, this implementation barrier inherently establishes CleanSE's practical advantage in resource-constrained environments.
\end{itemize}

\subsection{Real-world Dataset Test}\label{realworldsection}
The real-world test setting is different from that in section \ref{benchmark test}. We adopt a 3-MLP model as the baseline, all the datasets are run 5 trials (with 90\%/10\% train/test split). Since some of the datasets have limited instances, we did not test the predictive performance under different batch sizes as in benchmark experiments, consequently, we fixed the batch size as 64. For each trial, we empirically run about 250 epochs to make the model converge, and we also select the mean results of the last 10 epochs as the predictive performance.

The mean result and standard deviation are shown in Table \ref{real_test}. The best results are in bold, and the rank of each algorithm is shown in brackets in the table. From the observation, we can get the conclusion that:
\begin{itemize}
    \item The proposed CleanSE ranks 1st on 3/5 of the real world datasets, which exceeds all the other compared algorithms.
    \item On the \textbf{Lost} and \textbf{MSRCv2} dataset, CleanSE has weaker performance than on other datasets; this phenomenon is mainly due to the small dataset size and clean sample rate, which inhibit the proposed reweighting and countless schemes from extracting more accurate supervision information.  
    \item Here, the complementary label learning method \textbf{SCARE} underperforms significantly, mainly because it relies more on the non-candidate labels, which can further demonstrate that this kind of paradigm has internal disadvantages than PLL methods.
\end{itemize}
\begin{table}[htbp]\normalsize
\caption{Ablation Study of Knowledge Distillation}
\label{ablation-test}
\resizebox{\linewidth}{!}{
\begin{tabular}{cc|ccccc}
\hline\hline
\multicolumn{2}{c|}{Ablation}&\multicolumn{5}{c}{Ablation}\\ \cline{1-7}
reweight&countloss&\multicolumn{1}{c}{Lo}& \multicolumn{1}{c}{Bs}& \multicolumn{1}{c}{Ms}& \multicolumn{1}{c}{Sp}& \multicolumn{1}{c}{Yn}\\ \cline{1-7}
\XSolidBrush&\XSolidBrush&62.66\%&70.11\%&50.11\%&54.81\%&65.18\%\\ 
\Checkmark&\XSolidBrush&67.66\%&73.13\%&53.22\%&57.13\%&65.97\%\\
\XSolidBrush&\Checkmark&64.23\%&72.61\%&53.01\%&56.46\%&65.31\%\\
\Checkmark&\Checkmark& 69.33\%&75.21\%&55.05\%&58.90\%&66.97\%\\\hline\hline
\end{tabular}}
\end{table}
\subsection{Ablation Study}\label{alabtionsection}
The k-NN-based reweighting scheme and trackable count loss can fully leverage the clean samples from both local and global perspectives. To demonstrate the effectiveness of the two involved terms of our method, the ablation study is conducted on the 5 real-world datasets, \textbf{MSRCv2}, \textbf{Lost}, \textbf{BirdSong}, \textbf{Soccer Player}, and \textbf{Yahoo! News}. Table \ref{ablation-test} presents the results of the PLL model under different ablation scenarios.

The experimental data reveal that reweighting and count loss contribute to improving model training by utilizing clean samples. Noted that the reweighting loss can offer more guidance than count loss, this is because the global description of instance is a probabilistic estimation interval instead of an accurate distribution, and the reweighting strategy can effectively utilize every clean sample to eliminate the impact of false positive candidates, but we can still see that integration of the two scheme can positively boost the model to get good performance.

\subsection{Sensitive Tests}
In this experiment, two hyperparameters in relation to CleanSE need to be tested: temperature $T$ and regularization constraint hyperparameter $\lambda$.
\subsubsection{Temperature $T$}\label{temperature}
We conducted experiments on the real-world datasets; the experimental protocols mirrored the real-world evaluation setup described in section \ref{realworldsection}. Here, we fixed the hyperparameter $\lambda=0.001$, and the temperature $T$ is tested in the range $\{1,2,3,4,5\}$, which measures the enhanced degree of labels in the reweighting procedure.

\begin{figure}[tbph]
    \centering
    \includegraphics[width=0.45\textwidth]{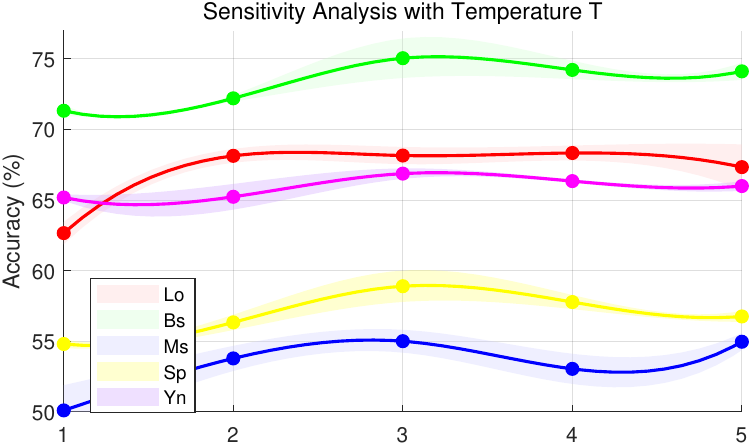}
    \caption{Sensitive test about Temperature $T$}
    \label{TemperatureT}
\end{figure}

Fig. \ref{TemperatureT} shows the visual result; As empirically demonstrated, the model reached top performance when $T=3$ in most cases, the results demonstrate that both high and low temperatures will affect the performance.
\subsubsection{Regularization Constraint Hyperparameter $\lambda$}
Except for the temperature $T$, we also test the impact of $\lambda$ on extracting the label distribution interval of PLL datasets. Similar to section \ref{temperature}, we fix the temperature $T$ as 3, then we record the predictive performance as $\lambda$ changes from \{1e-5,1e-4,...,1e-1,1\}.

\begin{figure}[tbph]
    \centering
    \includegraphics[width=0.45\textwidth]{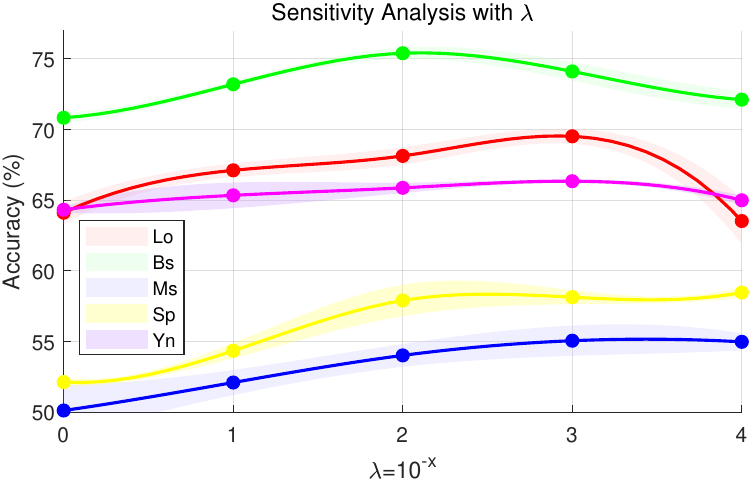}
    \caption{Sensitive test about regularization constraint hyperparameter $\lambda$}
    \label{Lambda}
\end{figure}

Fig. \ref{Lambda} quantitatively identifies the optimal $\lambda$ range as $[1e-4:1e-2]$, In practice, we suggest the balancing parameter to be 0.001.
\subsection{Statistic Tests}
As an essential part of evaluating new machine learning algorithms, statistic tests were used to examine the alignment between the hypothesis and predictive performance. In previous sections, the tests have multiple independent observations per cell. Thus, we consider employing the Friedman test \cite{test} to verify whether the average ranks of algorithms are significantly different under the null hypothesis. Noted that method number is $k=8$, and there are $N=25$ cases, so the statistics can be computed by the following equations:
\begin{equation}
    F_F=\frac{(N-1)\chi^2_F}{N(k-1)-\chi^2_F}\tag{9}
\end{equation}
 $\chi^2_F$ is derived by:
 \begin{equation}
    \chi^2_F=\frac{12N}{k(k+1)}(\sum_{i=1}^{k}R^2_i-\frac{k(k+1)^2}{4})\tag{10}
\end{equation}
Here, $R_i$ is the average rank of $i\text{-}th$ algorithm on all $N$ datasets, $R_i=\frac{1}{N}\sum_{i=1}^{N}r^j_i$. For \textbf{PL-CGR} and \textbf{DPCLS}, the ranks in the benchmark test are all settled as 7.5 out of 8 algorithms.
The statistics are shown in Table \ref{Clean_Friedman}, it can be observed that the null-hypothesis is rejected since the critical value $CD_f=2.06$, which means there is at least one algorithm that is significantly different from at least one other algorithm. To this end, we further conducted a Bonferroni-Dunn test \cite{test} as a post-hoc test.
 \begin{table}[thbp]
\centering
\caption{Friedman Test Statistics of 8 Algorithms (k=8), 25 Cases (N=25),$(\alpha=0.05)$}
\resizebox{\linewidth}{!}{
\begin{tabular}{lclc}
\hline \hline  Algorithm& Average Ranking & Algorithm &Average Ranking\\
\hline CC & 3.56 & 
RC & 3.00 \\
LWS & 3.16 & 
CAVL & 5.36  \\ 
SCARCE & 6.24 & 
PL-CGR & 6.52 \\
DPCLS & 7.08 & 
CleanSE & 1.12  \\\hline \hline
$F_F$ &69.4&Critical Value&2.06 \\
\hline \hline
\end{tabular}}
\label{Clean_Friedman}
\end{table}
\begin{figure}[tbph]
    \centering
    \includegraphics[width=0.45\textwidth]{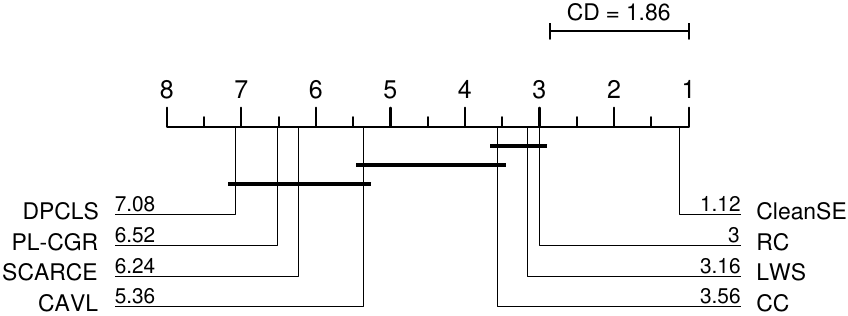}
    \caption{Bonferroni-Dunn test visualization}
    \label{BD-test}
\end{figure}

The CD diagram is presented in Fig. \ref{BD-test}. Based on the following equation:
\begin{equation}
CD=q_{\alpha}\sqrt{\frac{k(k+1)}{6N}}\notag
\end{equation}
The Bonferroni-Dunn test critical value is $CD_n=1.864$ at the significance of $\alpha=0.05$, here $q_{\alpha}=2.690$ according to \cite{test}. From the visual result in Fig \ref{BD-test}, we can observe that the proposed CleanSE shows a significant difference from the other 7 state-of-the-art PLL algorithms. Combine with the previous result, we can conclude that utilizing clean samples can help guide the model training in the correct direction. 
\subsection{Discussion}

In the benchmark test, the model performance shows a different trend when the batch size grows larger in the 4 datasets. Fundamentally rooted in their feature space characteristics. Three handwriting datasets exhibit sparse, clustered feature distributions. Conversely, that of \textbf{CIFAR-10} is full of extraneous features in every dimension; the excessively large embeddings will dilute primary features with smaller feature dimensions \cite{mixed}. Since the reweighting scheme is based on k-NN neighbours, the label distribution will be depicted more accurately when there are more samples within a batch on the three handwriting datasets, The reweighting classification loss took a dominant position since the count loss can roughly draw the distribution interval with more weak supervision information. When the feature space becomes dense, high-dimensional in the \textbf{CIFAR-10} dataset, the inessential feature can lead to a decline in performance, so when the batch size becomes bigger, k-NN has a weaker impact than countless.

As shown in Fig \ref{TemperatureT}, the accuracy rate has an upward trend after $T\geq4$ on the datasets \textbf{Lost} and \textbf{MSRCv2}. Recalling that these two datasets have only about 1000 instances with limited classes, and the clean sample rate is smaller compared with other datasets. Therefore, the most likely candidate label is relatively less reliable due to the structure of the datasets, resulting in the optimal temperature range to have a certain degree of randomness. Considering the last upward trend, we can't rule out the possibilities of better options than $T=3$, but in the range of $[0,5]$, we choose $T=3$ as the best temperature.
\section{Conclusion}
In this paper, we introduced a novel PLL approach called CleanSE. The key point is to utilize the concealed clean samples guide model training under normal PLL scenarios. Locally, we derived a k-NN-based reweighting method to disclose the relationship between clean and partial samples. Globally, we take advantage of the numbers of clean samples and partial samples to depict the label distribution interval.  Empirically, extensive experiments were conducted to show the superiority of CleanSE over 7 state-of-art PLL methods. The future work will be utilizing the clean samples in imbalanced PLL and noisy PLL scenarios. 



\bibliographystyle{elsarticle-num} 
\bibliography{sample.bib}

\begin{thebibliography}{10}
\expandafter\ifx\csname url\endcsname\relax
  \def\url#1{\texttt{#1}}\fi
\expandafter\ifx\csname urlprefix\endcsname\relax\def\urlprefix{URL }\fi
\expandafter\ifx\csname href\endcsname\relax
  \def\href#1#2{#2} \def\path#1{#1}\fi

\bibitem{videoannotation}
S.~Luo, S.~Jiang, D.~Cao, H.~Deng, J.~Wang, Z.~Qin, Weakly-supervised spatial--temporal video grounding via spatial--temporal annotation on a single frame, Knowledge-Based Systems 314 (2025) 113200.

\bibitem{privacy}
D.~Li, Z.~Yang, J.~Kang, M.~He, S.~Xie, Partially shared federated multiview learning, Knowledge-Based Systems 301 (2024) 112302.

\bibitem{PLLsurvey}
Y.~Tian, X.~Yu, S.~Fu, Partial label learning: Taxonomy, analysis and outlook, Neural Networks 161 (2023) 708--734.

\bibitem{binary}
T.~Cour, B.~Sapp, B.~Taskar, Learning from partial labels, The Journal of Machine Learning Research 12 (2011) 1501--1536.

\bibitem{dictionary}
Y.-C. Chen, V.~M. Patel, R.~Chellappa, P.~J. Phillips, Ambiguously labeled learning using dictionaries, IEEE Transactions on Information Forensics and Security 9~(12) (2014) 2076--2088.

\bibitem{graph}
G.~Lyu, S.~Feng, T.~Wang, C.~Lang, Y.~Li, Gm-pll: Graph matching based partial label learning, IEEE Transactions on Knowledge and Data Engineering 33~(2) (2019) 521--535.

\bibitem{progressive}
N.~Xu, B.~Liu, J.~Lv, C.~Qiao, X.~Geng, Progressive purification for instance-dependent partial label learning, in: International Conference on Machine Learning, PMLR, 2023, pp. 38551--38565.

\bibitem{robustness}
J.~Lv, B.~Liu, L.~Feng, N.~Xu, M.~Xu, B.~An, G.~Niu, X.~Geng, M.~Sugiyama, On the robustness of average losses for partial-label learning, IEEE Transactions on Pattern Analysis and Machine Intelligence 46~(5) (2023) 2569--2583.

\bibitem{CLSP}
S.~He, C.~Wang, G.~Yang, L.~Feng, Candidate label set pruning: A data-centric perspective for deep partial-label learning, in: The Twelfth International Conference on Learning Representations, 2023.

\bibitem{GLMNN-PLL}
X.~Gong, J.~Yang, D.~Yuan, W.~Bao, Generalized large margin $ k $ nn for partial label learning, IEEE Transactions on Multimedia 24 (2021) 1055--1066.

\bibitem{PLGP}
Y.~Zhou, J.~He, H.~Gu, Partial label learning via gaussian processes, IEEE transactions on cybernetics 47~(12) (2016) 4443--4450.

\bibitem{SP-PLL}
G.~Lyu, S.~Feng, T.~Wang, C.~Lang, A self-paced regularization framework for partial-label learning, IEEE Transactions on Cybernetics 52~(2) (2020) 899--911.

\bibitem{PL-PIE}
G.-Y. Lin, Z.-Y. Xiao, J.-T. Liu, B.-Z. Wang, K.-H. Liu, Q.-Q. Wu, Feature space and label space selection based on error-correcting output codes for partial label learning, Information Sciences 589 (2022) 341--359.

\bibitem{cc-rc}
L.~Feng, J.~Lv, B.~Han, M.~Xu, G.~Niu, X.~Geng, B.~An, M.~Sugiyama, Provably consistent partial-label learning, Advances in neural information processing systems 33 (2020) 10948--10960.

\bibitem{PRODEN}
J.~Lv, M.~Xu, L.~Feng, G.~Niu, X.~Geng, M.~Sugiyama, Progressive identification of true labels for partial-label learning, in: international conference on machine learning, PMLR, 2020, pp. 6500--6510.

\bibitem{KMT-PLL}
J.~Fan, L.~Huang, C.~Gong, Y.~You, M.~Gan, Z.~Wang, Kmt-pll: K-means cross-attention transformer for partial label learning, IEEE Transactions on Neural Networks and Learning Systems (2024).

\bibitem{pico+}
H.~Wang, R.~Xiao, Y.~Li, L.~Feng, G.~Niu, G.~Chen, J.~Zhao, Pico+: Contrastive label disambiguation for robust partial label learning, IEEE Transactions on Pattern Analysis and Machine Intelligence 46~(5) (2023) 3183--3198.

\bibitem{multilabelPLL}
A.~Tan, W.-Z. Wu, Partial multi-label learning via semi-supervised subspace collaboration, Knowledge-Based Systems 287 (2024) 111444.

\bibitem{multiviewPLL}
Z.~Wang, Y.~Xu, A two-stage multi-view partial multi-label learning for enhanced disambiguation, Knowledge-Based Systems 293 (2024) 111680.

\bibitem{semi-PLL}
A.~Tan, W.-Z. Wu, Partial multi-label learning via semi-supervised subspace collaboration, Knowledge-Based Systems 287 (2024) 111444.

\bibitem{imbalancedPLL}
H.~Wang, M.~Xia, Y.~Li, Y.~Mao, L.~Feng, G.~Chen, J.~Zhao, Solar: Sinkhorn label refinery for imbalanced partial-label learning, Advances in neural information processing systems 35 (2022) 8104--8117.

\bibitem{noisePLL}
M.~Xu, Z.~Lian, L.~Feng, B.~Liu, J.~Tao, Alim: adjusting label importance mechanism for noisy partial label learning, Advances in Neural Information Processing Systems 36 (2023) 38668--38684.

\bibitem{PLKD}
G.~Wang, J.~Huang, Y.~Lai, C.-M. Vong, Dealing with partial labels by knowledge distillation, Pattern Recognition 158 (2025) 110965.

\bibitem{KNNSAMPLE}
Z.~Zhu, Z.~Dong, Y.~Liu, Detecting corrupted labels without training a model to predict, in: International conference on machine learning, PMLR, 2022, pp. 27412--27427.

\bibitem{MLproof}
T.~Jo, Machine learning foundations, Supervised, Unsupervised, and Advanced Learning. Cham: Springer International Publishing 6~(3) (2021) 8--44.

\bibitem{simple}
K.~Ahmed, Z.~Zeng, M.~Niepert, G.~V.~d. Broeck, Simple: A gradient estimator for $ k $-subset sampling, arXiv preprint arXiv:2210.01941 (2022).

\bibitem{countloss}
V.~Shukla, Z.~Zeng, K.~Ahmed, G.~Van~den Broeck, A unified approach to count-based weakly supervised learning, Advances in Neural Information Processing Systems 36 (2023) 38709--38722.

\bibitem{mnist}
Y.~LeCun, L.~Bottou, Y.~Bengio, P.~Haffner, Gradient-based learning applied to document recognition, Proceedings of the IEEE 86~(11) (1998) 2278--2324.

\bibitem{fmnist}
H.~Xiao, K.~Rasul, R.~Vollgraf, Fashion-mnist: a novel image dataset for benchmarking machine learning algorithms, arXiv preprint arXiv:1708.07747 (2017).

\bibitem{kmnist}
T.~Clanuwat, M.~Bober-Irizar, A.~Kitamoto, A.~Lamb, K.~Yamamoto, D.~Ha, Deep learning for classical japanese literature, arXiv preprint arXiv:1812.01718 (2018).

\bibitem{CIFAR10}
A.~Krizhevsky, G.~Hinton, et~al., Learning multiple layers of features from tiny images (2009).

\bibitem{lost}
T.~Cour, B.~Sapp, B.~Taskar, Learning from partial labels, The Journal of Machine Learning Research 12 (2011) 1501--1536.

\bibitem{birdsong}
F.~Briggs, X.~Z. Fern, R.~Raich, Rank-loss support instance machines for miml instance annotation, in: Proceedings of the 18th ACM SIGKDD International Conference on Knowledge Discovery and Data Mining, 2012, pp. 534--542.

\bibitem{msrcv2}
L.~Liu, T.~Dietterich, A conditional multinomial mixture model for superset label learning, Advances in Neural Information Processing Systems 25 (2012).

\bibitem{soccerplayer}
Z.~Zeng, S.~Xiao, K.~Jia, T.-H. Chan, S.~Gao, D.~Xu, Y.~Ma, Learning by associating ambiguously labeled images, in: Proceedings of the IEEE/CVF Conference on Computer Vision and Pattern Recognition, 2013, pp. 708--715.

\bibitem{yahoonews}
M.~Guillaumin, J.~Verbeek, C.~Schmid, Multiple instance metric learning from automatically labeled bags of faces, in: Computer Vision--ECCV 2010: 11th European Conference on Computer Vision, Heraklion, Crete, Greece, September 5-11, 2010, Proceedings, Part I 11, Springer, 2010, pp. 634--647.

\bibitem{lws}
H.~Wen, J.~Cui, H.~Hang, J.~Liu, Y.~Wang, Z.~Lin, Leveraged weighted loss for partial label learning, in: International conference on machine learning, PMLR, 2021, pp. 11091--11100.

\bibitem{cavl}
F.~Zhang, L.~Feng, B.~Han, T.~Liu, G.~Niu, T.~Qin, M.~Sugiyama, Exploiting class activation value for partial-label learning, in: International conference on learning representations, 2021.

\bibitem{SCARCE}
W.~Wang, T.~Ishida, Y.-J. Zhang, G.~Niu, M.~Sugiyama, Learning with complementary labels revisited: The selected-completely-at-random setting is more practical, in: International Conference on Machine Learning, PMLR, 2024, pp. 50683--50710.

\bibitem{PL-CGR}
Z.~Zhang, Z.~Liu, H.~Lu, Partial label learning via cost-guided retraining, in: ECAI 2024, IOS Press, 2024, pp. 2170--2177.

\bibitem{DPCLS}
Y.~Jia, F.~Yang, Y.~Dong, Partial label learning with dissimilarity propagation guided candidate label shrinkage, Advances in neural information processing systems 36 (2023) 34190--34200.

\bibitem{test}
J.~Dem{\v{s}}ar, Statistical comparisons of classifiers over multiple data sets, The Journal of Machine Learning Research 7 (2006) 1--30.

\bibitem{mixed}
F.~Yang, J.~Cheng, H.~Liu, Y.~Dong, Y.~Jia, J.~Hou, Mixed blessing: Class-wise embedding guided instance-dependent partial label learning, arXiv preprint arXiv:2412.05029 (2024).

\end{thebibliography}





\end{document}